\documentclass[conference]{IEEEtran}
\IEEEoverridecommandlockouts
\usepackage{cite}
\usepackage{amsmath,amssymb,amsfonts}
\usepackage{algorithmic}
\usepackage{graphicx}
\usepackage{textcomp}
\usepackage{xcolor}
\def\BibTeX{{\rm B\kern-.05em{\sc i\kern-.025em b}\kern-.08em
    T\kern-.1667em\lower.7ex\hbox{E}\kern-.125emX}}
\usepackage{multirow}
\usepackage{cite}    
\makeatletter
\let\NAT@parse\undefined
\makeatother
\usepackage[colorlinks,linkcolor=blue,citecolor=blue]{hyperref}
\usepackage{threeparttable}
\usepackage{array}
\usepackage{colortbl}  % 如果要使用表格的颜色
\definecolor{tab_others}{RGB}{235, 235, 235}
\usepackage{rotating}
\def \pzo {\phantom{0}}
\usepackage{xcolor}
\newcommand{\red}[1]{\textcolor{red}{#1}}
\newcommand{\blue}[1]{\textcolor{blue}{#1}}
\def \pzo {\phantom{0}}
\usepackage{booktabs}
\begin{document}

\title{HSS-IAD: A Heterogeneous Same-Sort Industrial Anomaly Detection Dataset}

% \author{Anonymous ICME submission}
\author{
    \IEEEauthorblockN{Qishan Wang\textsuperscript{\rm a}, Shuyong Gao\textsuperscript{\rm b,*}, Junjie Hu\textsuperscript{\rm b}, Jiawen Yu\textsuperscript{\rm a}, Xuan Tong\textsuperscript{\rm a}, You Li\textsuperscript{\rm a}, Wenqiang Zhang\textsuperscript{a,b,*}
    \thanks{*Corresponding author. This work was supported by 
    the National Natural Science Foundation of China under Grant 62072112,
    the Hexi University President’s Fund for Young Scientists Research Project under Grant QN202204,
    and the Gansu Provincial Education Scientific and Technological Innovation Project under Grant 2023A-130.
    }} 
    \IEEEauthorblockA{\textsuperscript{\rm a} Academy for Engineering and Technology, Fudan University, Shanghai, China}
    \IEEEauthorblockA{\textsuperscript{\rm b} School of Computer Science, Fudan University, Shanghai, China}
    \IEEEauthorblockA{\{qswang20, sygao18, wqzhang\}@fudan.edu.cn}
}

\maketitle

\begin{abstract}
Multi-class Unsupervised Anomaly Detection algorithms (MUAD) are receiving increasing attention due to their relatively low deployment costs and improved training efficiency. However, the real-world effectiveness of MUAD methods is questioned due to limitations in current Industrial Anomaly Detection (IAD) datasets. These datasets contain numerous classes that are unlikely to be produced by the same factory and fail to cover multiple structures or appearances. Additionally, the defects do not reflect real-world characteristics.
Therefore, we introduce the Heterogeneous Same-Sort Industrial Anomaly Detection (HSS-IAD) dataset, which contains 8,580 images of metallic-like industrial parts and precise anomaly annotations. These parts exhibit variations in structure and appearance, with subtle defects that closely resemble the base materials. We also provide foreground images for synthetic anomaly generation. Finally, we evaluate popular IAD methods on this dataset under multi-class and class-separated settings, demonstrating its potential to bridge the gap between existing datasets and real factory conditions. The dataset is available at \href{https://github.com/Qiqigeww/HSS-IAD-Dataset}{https://github.com/Qiqigeww/HSS-IAD-Dataset}.
\end{abstract}

\section{Introduction}
\label{sec:intro}

\begin{figure}[tbp]
\centerline{\includegraphics[width=0.78\linewidth]{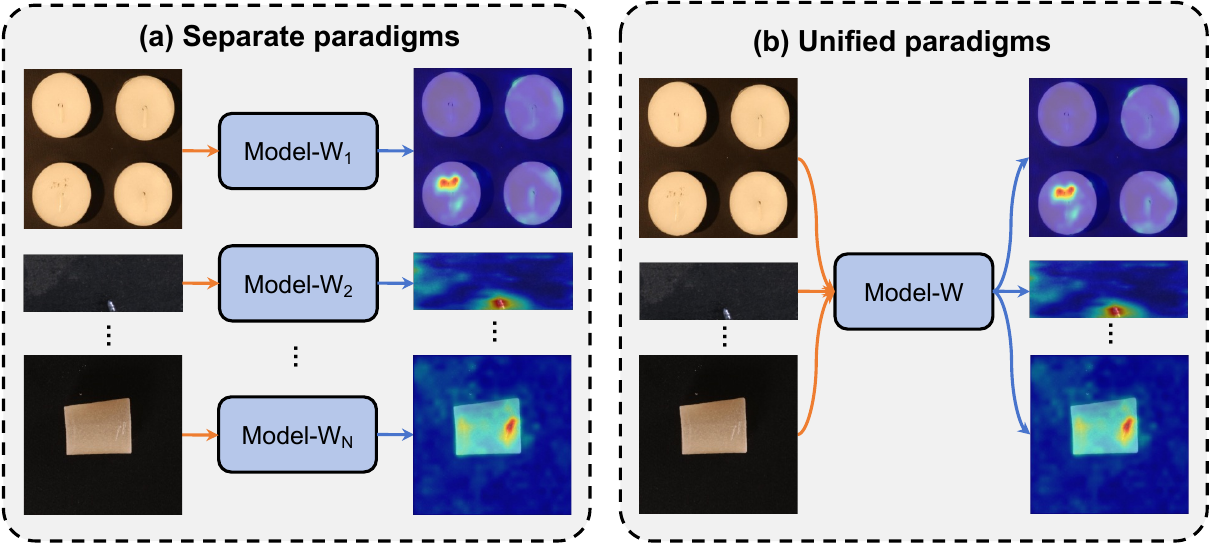}}
\caption{Comparison of training paradigms.}
\label{train_para}
\end{figure}

Industrial Anomaly Detection (IAD) aims to identify and localize unknown patterns in product images that significantly deviate from the norm. This approach requires only normal samples rather than a large number of defective samples and their labels, which can potentially enhance the performance and efficiency of defect detection in casting surfaces~\cite{wang2024casting} and aluminum profiles.
% ~\cite{liu2021semi}.
Most existing methods train separate models for each object class, as illustrated in Fig.~\ref{train_para}(a). However, such a one-class-one-model setting leads to substantial training and storage costs. Moreover, it is unsuitable for real-world scenarios where multiple product categories need to be efficiently detected within the same factory. Recently, UniAD~\cite{you2022unified} and subsequent studies~\cite{zhang2023vit} 
% ~\cite{he2024diad,zhang2023vit} 
have proposed training a unified model for Multiclass Unsupervised Anomaly Detection (MUAD), as depicted in Fig.~\ref{train_para}(b). However, developing a model that captures the distribution across multiple classes remains a considerable challenge.

\begin{figure}[tbp]
\centerline{\includegraphics[width=1.0\linewidth]{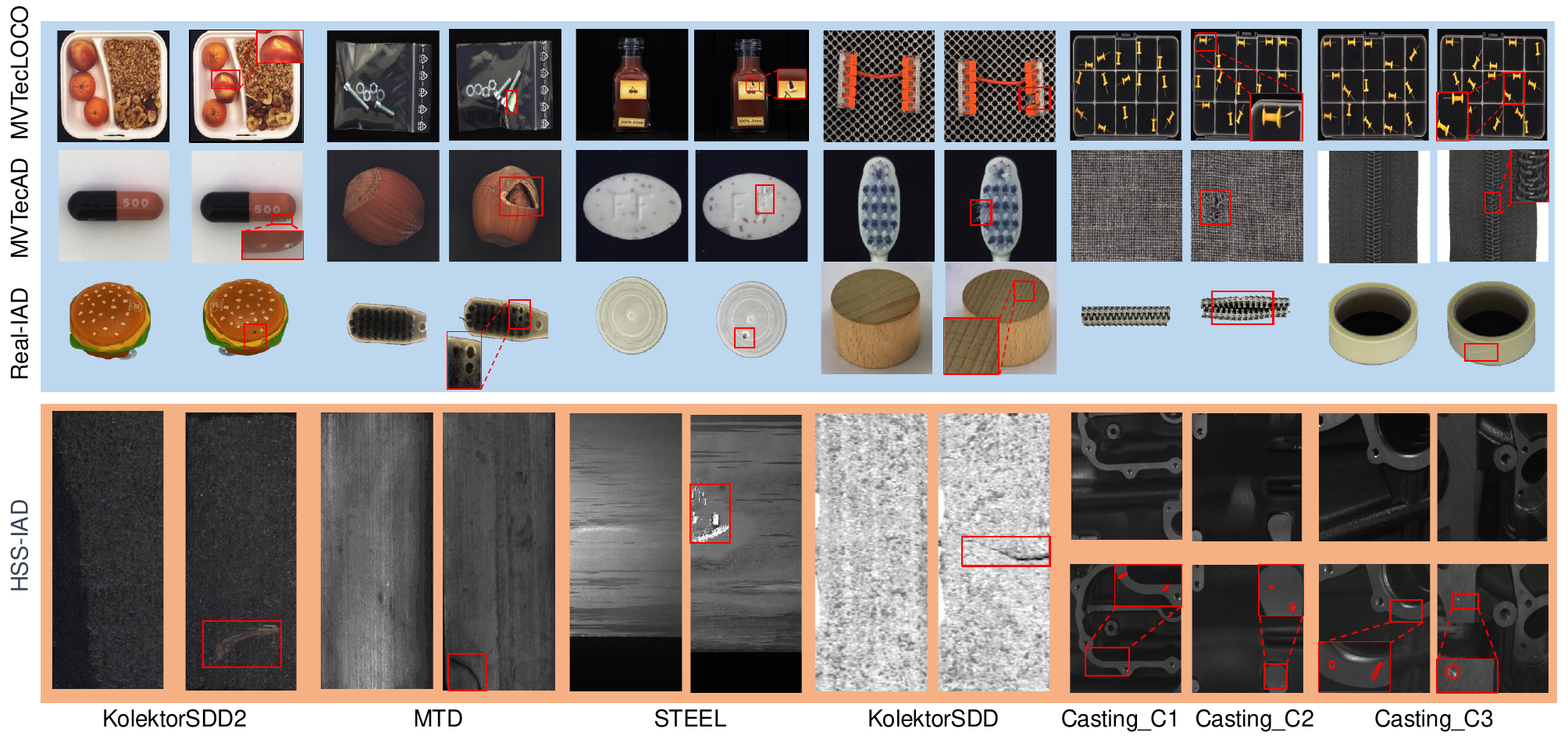}}
\caption{Normal and anomalous samples from existing datasets and the HSS-IAD dataset. Close-ups of anomalies are highlighted in red boxes. Compared to existing datasets (e.g., MVTecAD, RealIAD), which include diverse classes like fruits, snacks, pills, etc., the HSS-IAD dataset has the following characteristics: 1.) The subclasses consist of same sort industrial products while exhibiting variations in structure or appearance. 2.) The subtle and variable defects closely resemble the material itself.}
\label{fig_data}
\end{figure}

Under the MUAD setting, current common datasets include MVTecAD~\cite{bergmann2019mvtec}, Real-IAD~\cite{wang2024real}, and MVTec LOCO~\cite{bergmann2022MVTec_LOCO}, which consist of various categories, such as fruits, snacks, pills, wine bottles, and fabrics, as shown in Fig.~\ref{fig_data}. These datasets have stimulated academic interest in industrial anomaly detection research and has given birth to many innovative methods. However, they overlook an important fact: products produced by the same plant should be of the same sort but diverse structures or appearances. 
For example, an automaker manufactures several types of workpieces, such as steel plates with different carbon content, castings, and magnetic tiles, but does not produce fruits or toothbrushes~\cite{survey_ad}. Additionally, real industrial defects are often very subtle and similar to the material, easily confused with processing marks, oil stains, etc. The settings of these datasets do not consider the real working conditions of the factory. Consequently, it is difficult to conclude that current advanced MUAD methods trained on these datasets can still maintain high accuracy under such challenging conditions.

\begin{figure*}[htbp]
\centerline{\includegraphics[width=0.94\linewidth]{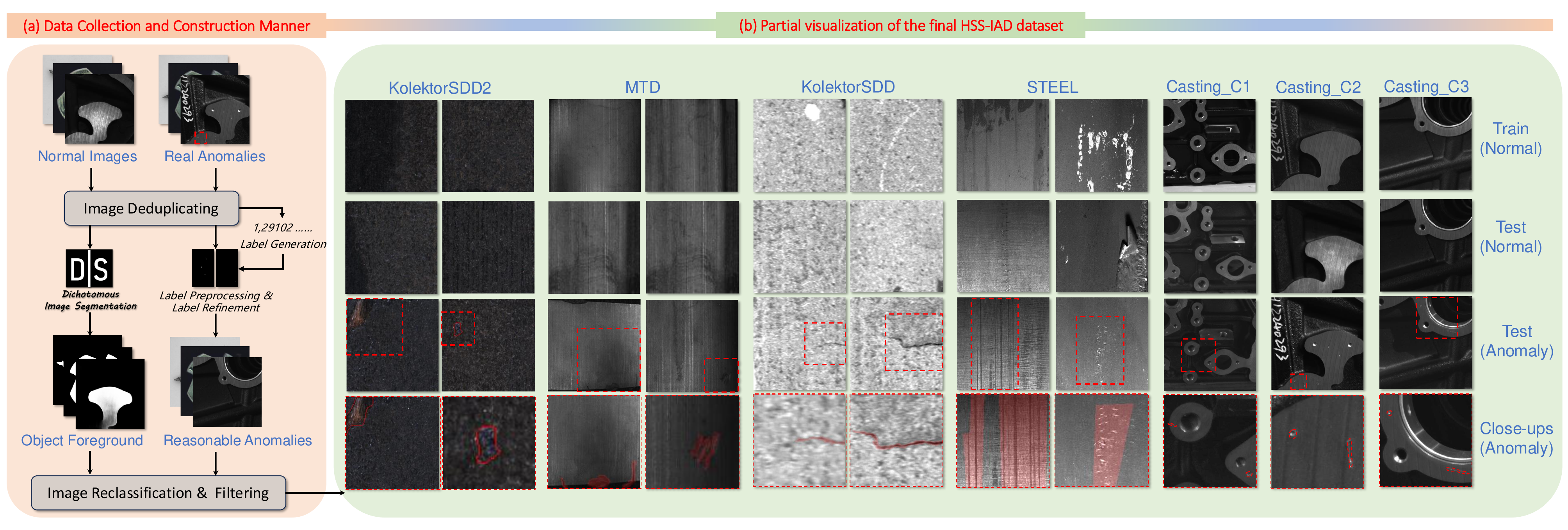}}
\caption{Data collection pipeline for our proposed HSS-IAD dataset. (a) The collection procedure encompasses various stages, including data collection, deduplication, foreground generation, label generation, label refinement, and reclassification and filtering.
(b) The samples in the HSS-IAD dataset are divided into the training set and the test set. The training set consists solely of normal samples of industrial products made of metal or magnetic tile. These products have same sort but differ significantly in structure or appearance. The test set consists of normal samples and abnormal samples with subtle and variable defects. Close-up figures of abnormal samples, with highlighted abnormal areas, are displayed in the last row along with their pixel-precise ground truth labels.}
\label{fig_dataset}
\end{figure*}

To make the dataset closer to real application scenarios and address the limitations of existing datasets, we introduce a new benchmark, HSS-IAD, collected from same sort industrial product while exhibiting variations in structure or appearance in this paper. 
The benchmark comprises seven categories, including both object and texture types.
Defects include subtle scratches, cold cracks, minor spots, etc., while backgrounds interference sources like oxide scales, machining marks, and oil stains. The workpiece categories include electrical commutators, magnetic tiles, flat sheet steel and engine castings, totaling 8,580 images.
In addition, given that data augmentation-based methods~\cite{zhang2023destseg, zavrtanik2021draem} have achieved superior detection performance in IAD, we also employ the DIS method~\cite{qin2022DIS} to provide the foreground images of objects, facilitating the generation of a large number of synthetic anomalies.
Besides, we conducted a series of benchmark tests using state-of-the-art (SoTA) unsupervised anomaly detection algorithms on the new dataset to assess their performance.
The experimental results show that while these algorithms perform well on the original datasets, there is still considerable room for improvement on the new dataset, highlighting its challenge and potential to drive algorithmic improvement and innovation.

In summary, the main contributions of this paper are as follows.
\begin{itemize}
    \item To bridge the gap between the actual application of MUAD and the available datasets, we constructed the Heterogeneous Same-Sort Industrial Anomaly Detection (HSS-IAD) Dataset. It takes into account various practical factors such as product sort similarity, structural or appearance variations and the subtle, easily confusable nature of defects. The dataset closely aligns with real-world needs for detecting anomalies in heterogeneous products of the same sort.
    \item We report the performance of popular anomaly detection methods on the HSS-IAD dataset under both class-Separated and multi-Class settings, providing a highly challenging benchmark to promote the development of IAD algorithms for industrial scenarios.
\end{itemize}

\section{Background}

A large number of IAD datasets have been used in academic research. Early industrial anomaly detection methods were applied on KolektorSDD~\cite{tabernik2020KolektorSDD} and KolektorSDD2~\cite{bovzivc2021ksdd2}, which contains real industrial images, but the single category limits the generalization ability of the algorithms. 
Subsequently, MTD~\cite{huang2020MTD}, STEEL~\cite{severstal-steel-defect-detectionSTEEL}, Casting~\cite{wang2024casting}, and MPDD~\cite{jezek2021mpdd} were proposed. MTD includes five types of anomalies, and STEEL refers to the Severstal Steel defect dataset used in the Kaggle competition. However, both of these datasets exhibit minimal appearance variation and lack structural diversity. Casting contains six types of structural surface, but it lacks parts with changes in appearance alone.
MPDD specifically focused on the issue of defect detection during painted metal parts fabrication, without addressing practical challenges such as the high similarity between defects and process features.
BTAD includes three products, but the number of images is relatively small.xs
With the emergence of MVTecAD~\cite{bergmann2019mvtec} and VisA~\cite{zou2022visa}, many innovative methods have come to the fore. MVTecAD contains 5,354 high-resolution images of ten objects and five texture categories. Each image contains only one centered object, and the defects are also apparent. VisA covers 12 objects across three domains and presents several challenges, such as objects with complex structures, multiple instances located differently in a single view, etc.
The semantics of objects from the same domain vary significantly, ranging from capsules to candles. Many product categories are unlikely to be produced or updated concurrently within the same factory, making it difficult to meet the real detection needs of factories. Later, MVTec LOCO~\cite{bergmann2022MVTec_LOCO} explicitly focused on the detection of both structural and logical anomalies, requiring models to understand the underlying logical or geometrical relationships in anomaly-free data. 
Recently, Wang et al.~\cite{wang2024real} proposed a large-scale, real-world, and multi-view Real-IAD dataset, featuring 150K high-resolution images spanning 30 different objects.
% to bring it closer to real application scenarios. 
% This dataset contains 150K high-resolution images of 30 different objects. 
However, there is still a significant gap between the current dataset setup and practical application scenarios.

To date, existing visual IAD datasets have not focused on detecting subtle and confusable defects in parts of the same sort but with significantly different structures or appearances.
To fill this gap, we introduce the Heterogeneous Same-Sort Industrial Anomaly Detection (HSS-IAD) Dataset, which covers various materials, structures, and defect types found in processing workshops.

\section{PROPOSED DATASET}\label{dataset}
\begin{figure*}[htbp]
\centerline{\includegraphics[width=1.0\textwidth]{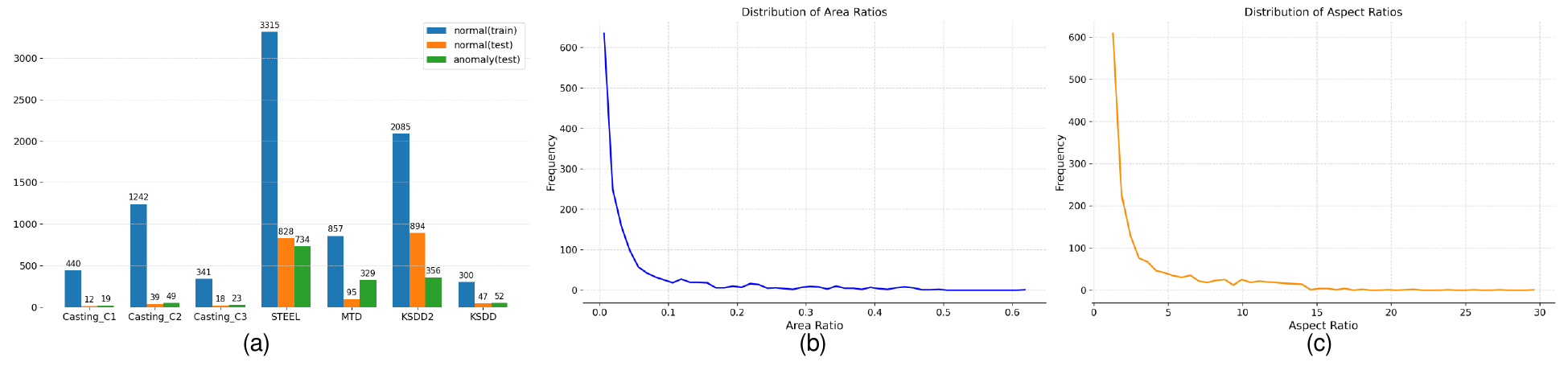}}
\caption{Statistical information of our proposed HSS-IAD dataset: (a) Distribution of anomaly/normal image quantities in training and test across different categories. (b) Statistics of the percentage of the image area occupied by anomaly region. (c) Statistics of the aspect ratio of the minimum bounding rectangles of the defect.}
\label{fig_statistical}
\end{figure*}
\begin{figure}[htbp]
\centerline{\includegraphics[width=0.8\linewidth]{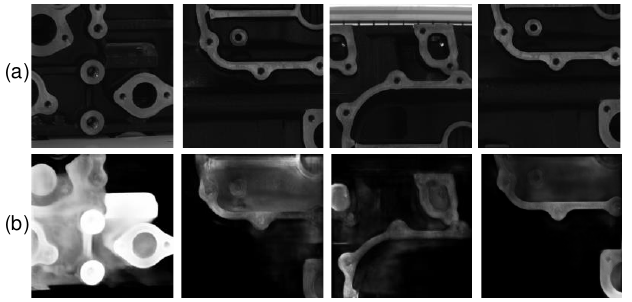}}
\caption{(a) Casting image; (b) Foreground image of Casting.}
\label{DIS_Seg}
\end{figure}
\begin{table}[tbp]    
    \caption{Comparison of Current Popular Datasets and HSS-IAD on Different Attributes: Category (Number of Metallic-like Product Categories), Samples (Number of Samples), Proportion (Proportion of Abnormal Pixels), and Similarity (Similarity Between Defect and Background).}
    \begin{center}
    \begin{tabular}{|c|c|c|c|c|}
        \hline
        Dataset & Category & Samples & Proportion & Similarity \\ 
        \hline
        MVTecAD & 3 & 1157 & 5.4\% & Low \\
        MVTec LOCO & 2 & 1512 & 5.4\% & Low \\
        Real-IAD & 4 & N & N & Low \\
        \hline
        HSS-IAD & \textbf{7} & \textbf{8580} & \textbf{3.0\%} & \textbf{High} \\
        \hline
        % \multicolumn{5}{p{7cm}}{\raggedright Category: number of metallic-like Product Categories.\\ Proportion: proportion of abnormal pixel.\\ Similarity: the similarity between defect and background.}
        % \footnotesize
    \end{tabular}
    \label{tab:datasetcompare}
    \end{center}
\end{table}
\subsection{Data Collection and Construction Manner}
To obtain images of same sort industrial products while exhibiting variations in structure or appearance, we first \textbf{selected} and classified samples from the KolektorSDD2~\cite{bovzivc2021ksdd2}, Casting~\cite{wang2024casting}, MTD~\cite{huang2020MTD}, KolektorSDD~\cite{tabernik2020KolektorSDD} and STEEL~\cite{severstal-steel-defect-detectionSTEEL} datasets.

Next, we carefully evaluated the quality of the images, \textbf{removing blurry, duplicate}, and low-quality images, as shown in Fig. \ref{fig_dataset} (a). Considering that data augmentation-based methods have proven highly effective in IAD task, we provided the \textbf{foreground images} of objects in image using the DIS method~\cite{qin2022DIS} to facilitate generating more diverse and realistic synthetic anomalies during the training phase, as shown in Fig~\ref{DIS_Seg}. Since the STEEL dataset does not provide ground truth for anomalies, we utilized textual documents containing defect area bounding box information to \textbf{generate pixel-level masks} for the corresponding images. Upon inspection, it was found that some annotated anomaly regions were either too large and did not match the actual anomalies (e.g., in the STEEL dataset) or were too small, causing the anomalies to disappear after label preprocessing (e.g., in the MTD dataset). Additionally, misclassified test samples were corrected. Therefore, we performed \textbf{zero image check and manual label filtering}, eliminating samples with large annotation errors or tiny detection areas.

Finally, we performed \textbf{image reclassification and filtering}. For instance, the current six-class setup of the casting dataset~\cite{wang2024casting} lacked meaningful distinctions. Many samples had excessive background noise or interference, which provided limited value for model training. To address this issue, we reselected a subset of casting samples and divided them into the following three more distinct categories: (1) The images should include both cast and machined surfaces, with the machined surfaces being relatively simple, lacking threaded holes but potentially showing oil stains or machining marks; (2) The machined surfaces should feature more complex structures, such as multiple threaded holes and other process features that could act as sources of interference; (3) The samples should exhibit more complicated interferences, where the machined surfaces have few or no threaded holes. Additionally, we selected a number of normal samples with confusing attached elements that could be mistaken for anomalies. Examples of these are found in the KolektorSDD~\cite{tabernik2020KolektorSDD} and STEEL~\cite{severstal-steel-defect-detectionSTEEL} datasets, as illustrated in Fig. \ref{fig_dataset}. To reflect the detection challenges in real scenarios, we selected the test set to include subtle and varied defects that are easily confused with surface machining marks, oil stains, machining residues, process features, etc., as shown in the abnormal samples in Fig.~\ref{fig_dataset}.
% More comprehensive visualizations of the dataset samples are provided in Fig.~\ref{fig:vis_hss-iad} of the Supplementary Material.

\subsection{Dataset Statistics}
As shown in Table~\ref{tab:datasetcompare}, compared to mainstream datasets, HSS-IAD significantly increases both the metallic-like product categories and samples, totaling 8,580.
Furthermore, the anomalous pixel ratio in HSS-IAD is 3.0\%, indicating its suitability for tackling the complex challenges of small defect detection in industrial settings. Unlike other datasets, the similarity between defects and the background in our dataset is notably high, which increases the difficulty of detecting anomalies.

Fig.~\ref{fig_statistical} presents statistics of our HSS-IAD dataset. Fig.~\ref{fig_statistical} (a) shows the number of normal and abnormal data in training and testing across different categories. HSS-IAD covers multiple metallic-like categories and provides richer real-world industrial scenarios. The sample numbers for the KSDD2, Casting, and STEEL datasets are considerably larger than those of other categories, as they are the most widely used materials in industrial applications. The proportion of defect areas (Fig.~\ref{fig_statistical} (b)) and the range of defect ratios for each class(Fig.~\ref{fig_statistical} (c)) are larger, indicating a higher difficulty level of the dataset, as also demonstrated by the experiments in Table~\ref{tab:miad}. 
A greater challenge can better differentiate the performance of various methods. These characteristics make the dataset closer to real-world scenarios and help address the limitations of existing datasets.

\section{BENCHMARK}
We establish two evaluation protocols, including class-separated and multi-class industrial anomaly detection settings for all 7 categories in the HSS-IAD dataset.
\begin{table}[tbp]
    \centering
    \caption{Performance Comparisons on HSS-IAD and Existing IAD Datasets under multi-class IAD Setting (\%). The results are averaged over all categories on each dataset. We report the averaged value and standard deviation (denoted as ``mean$\pm$std") based on the results of different methods for each metric. The lower the value means the corresponding anomaly detection task is more challenging.}
    \renewcommand{\arraystretch}{1.2}
    \resizebox{\linewidth}{!}{
        \begin{tabular}{l|l|cc|cc|cc|c}
            \hline
            &  & \multicolumn{2}{c|}{Embedding-based}  & \multicolumn{2}{c|}{Data-Aug-based} & \multicolumn{2}{c|}{Reconstruction-based}  &\multirow{2}{*}{\textbf{Mean$\pm$Std}} \\ 
             \multirow{-2}{*}{\textbf{Datasets}}     & \multirow{-2}{*}{\textbf{Metric$\uparrow$}} 
            % & \textbf{DeSTSeg\cite{zhang2023destseg}}   &\textbf{SimpleNet\cite{liu2023simplenet}} & \textbf{DRÆM\cite{zavrtanik2021draem}} & \textbf{UniAD\cite{you2022unified}}  & \textbf{RD4AD\cite{deng2022rd4ad}} & \textbf{Dinomaly\cite{guo2024dinomaly}}   &
            & \textbf{DeSTSeg}   &\textbf{SimpleNet} & \textbf{DRÆM} & \textbf{UniAD}  & \textbf{RD4AD} & \textbf{Dinomaly}   &          
            \\ 
            \hline
            \rowcolor{blue!6}  \cellcolor{white}{} & I-AUROC   &89.2	&95.3	&88.8	&96.5	&94.6	&99.6	&94.0	$\pm$    \pzo3.9\\
            \multirow{-2}{*}{MVTec-AD~\cite{bergmann2019mvtec}}                          
            & P-AUPRO   &64.8	&86.5	&71.1	&90.7	&91.1	&94.8	&83.2	$\pm$  11.2\\
            \hline
            \rowcolor{blue!6}  \cellcolor{white}{} & I-AUROC   &88.9	&87.2	&79.5	&88.8	&92.4	&98.7	&89.3	$\pm$    \pzo5.8\\
            \multirow{-2}{*}{VisA~\cite{zou2022visa}}                          
            & P-AUPRO   &67.4	&81.4	&59.1	&85.5	&91.8	&94.5	& 80.0	$\pm$   12.8\\
            \hline
            \rowcolor{blue!6}  \cellcolor{white}{} & I-AUROC   &82.3	&57.2	&---	&83.0	&82.4	&89.3	&78.8	$\pm$ 11.1\\
            \multirow{-2}{*}{Real-IAD~\cite{wang2024real}}                          
            & P-AUPRO    &40.6	&39.0	&---	&86.7	&89.6	&93.9	&70.0	$\pm$   24.7\\  
            % \hline
            % \rowcolor{blue!6}  \cellcolor{white}{} & I-AUROC  &73.6	&54.3	&63.7	&63.4	&69.2	&77.7	&67.0	$\pm$ \pzo7.6\\
            % \multirow{-2}{*}{HSS-IAD}                          
            % & P-AUPRO  &55.6	&20.5	&24.1	&49.6	&58.1	&54.8	&43.8	$\pm$    15.4\\
            % \hline
%
            \hline
            \rowcolor{blue!6}  \cellcolor{white}{} & I-AUROC  &93.5	&94.0	&69.3	&94.5	&94.1	&95.4	&90.1 $\pm$\pzo9.3\\
            \multirow{-2}{*}{BTAD~\cite{mishra2021btad}}                          
            & P-AUPRO  &72.9	&69.6	&16.1	&78.9	&79.9	&76.5	&65.6    $\pm$   22.4\\
            \hline
            \rowcolor{blue!6}  \cellcolor{white}{} & I-AUROC  &92.6	&90.6	&60.2	&80.1	&90.3	&97.2	&85.2	$\pm$   12.3\\
            \multirow{-2}{*}{MPDD~\cite{jezek2021mpdd}}                          
            & P-AUPRO  &78.3	&90.0	&21.8    &83.8	&95.2	&96.6	&77.6	$\pm$   25.7\\
            \hline
            \rowcolor{blue!6}  \cellcolor{white}{} & I-AUROC  &73.6	&54.3	&63.7	&63.4	&69.2	&77.7	&67.0	$\pm$\pzo7.6\\
            \multirow{-2}{*}{HSS-IAD}                          
            & P-AUPRO  &55.6	&20.5	&24.1	&49.6	&58.1	&54.8	&43.8	$\pm$    15.4\\
            \hline
        \end{tabular}
    }
    \vspace{-1.5em}
    \label{tab:compare_other_datasets}
\end{table}

\begin{table*}[tbp]
    \centering
    \caption{Performance (I-AUROC/P-AP/P-AUPRO) Comparisons on HSS-IAD under multi-class IAD Setting (\%). $\dag$: method designed for MUAD. The best results are marked in red and the second-best ones are marked in blue.}
    \renewcommand{\arraystretch}{1.2}
    \resizebox{0.95\linewidth}{!}{
        {\scriptsize
        \begin{tabular}{l| cc | cc |cc}
        \hline
        \multirow{2}{*}{\textbf{Type}} & \multicolumn{2}{c|}{Embedding-based}  & \multicolumn{2}{c|}{Data-Aug-based} & \multicolumn{2}{c}{Reconstruction-based} \\ 
        & \textbf{DeSTSeg\cite{zhang2023destseg}}   &\textbf{SimpleNet\cite{liu2023simplenet}} & \textbf{DRÆM\cite{zavrtanik2021draem}} & \textbf{UniAD\cite{you2022unified}}$\dag$  & \textbf{RD4AD\cite{deng2022rd4ad}} & \textbf{Dinomaly\cite{guo2024dinomaly}}$\dag$ \\
        \hline         % \blue{91.3}        \red{91.4}        
        MTD  &71.2	/ \blue{26.4}	/ \blue{70.6}	&60.3	/    11.0	/ 20.5	&50.7	/ 11.7	/ 20.6	&80.7	/ 21.4	/ \red{74.4}	&\blue{86.2}	/ 24.5	/ 69.2	& \red{93.6}	/ \red{38.4}	/ 62.2	\\
        
        STEEL &59.1	/ 17.9	/ 36.0	&52.5	/ \pzo3.9 / \pzo9.6	&\blue{62.9}	/ 13.8	/ 18.3	&50.6	/ \blue{19.3}	/ 37.6	&51.0	/ 18.6	/ \red{48.9}	&\red{63.0}	/ \red{30.7}	/ \blue{43.3} \\ 
        
        KolektorSDD &80.8	/ \red{17.5}	/ 33.8		&60.1 / \pzo2.3 / 22.2 &52.5	/ \pzo0.2	/ \pzo6.0 & 66.1 / \pzo4.5 / 29.6		&\blue{80.9} / 12.2 / \blue{66.1} 	&\red{93.0}	/ \blue{14.8} / \red{90.5}	\\ 
        
        KolektorSDD2 &\red{95.1}	/ \red{69.5}	/ 91.8	        	&72.0 / 49.0 / 78.7	&76.5	/ 29.7	/ 46.2	& 94.7	/ 46.2 / \red{94.1}	&\blue{94.8} / 52.0 / \blue{92.6}  &93.3 / \blue{68.3} / 87.6	\\
        
        Casting\_C1 &   \blue{69.7}	/ \red{\pzo4.3}	/ \red{49.4}    	&45.6 / \pzo0.1 / \pzo2.7 	&66.2	/ \pzo1.0	/ 22.8	& 50.4	/ \pzo0.9 / 31.5	&51.3	/ \blue{\pzo3.6}	/ \blue{43.3}    &\red{71.5}	/ \pzo2.6	/ 40.2	\\
        
        Casting\_C2 &   \blue{63.0}	/ \pzo1.5	/ \blue{54.8}    	&55.2 / \pzo0.1 / \pzo1.4 	&  \red{67.3}	/ \pzo0.6	/ 29.2	& 50.5	/ \pzo1.2 / 47.8	&56.7	/ \blue{\pzo2.0}	/ \red{56.1}		&61.6	/ \red{\pzo2.1}	/ 45.2	\\
        
        Casting\_C3 &   \red{76.3}	/ \red{\pzo1.2}	/ \red{53.1}    	& 34.5 / \pzo0.1 / \pzo8.5 	&  \blue{69.6}	/ \pzo0.3	/ 25.7	& 50.5	/ \pzo0.4 / \blue{32.1}	&63.3	/ \blue{\pzo1.0}	/ 30.2		&67.6	/ \pzo0.4	/ 14.9	\\        
        \hline
        
        Average &   \blue{73.6}	/  \blue{19.8}	/  \blue{55.6} 	& 54.3 /  \pzo9.5 /   20.5 	&   63.7	/ \pzo8.2	/ 24.1	&     63.4 / 13.4 / 49.6	&  69.2 / 16.3 / \red{58.1}  &    \red{77.7}	/ \red{22.5}	/ 54.8   \\        
        \hline
        \end{tabular}
        }
    }
    \vspace{-1.5em}
    \label{tab:miad}
\end{table*}

\begin{table*}[tbp]
    \centering
    \caption{Performance (I-AUROC/P-AP/P-AUPRO) Comparisons on HSS-IAD under class-separated IAD Setting (\%).}
    \renewcommand{\arraystretch}{1.2}
    \resizebox{0.95\linewidth}{!}{
        {\scriptsize
        \begin{tabular}{l| cc | cc |cc}
        \hline
        \multirow{2}{*}{\textbf{Type}} & \multicolumn{2}{c|}{Embedding-based}  & \multicolumn{2}{c|}{Data-Aug-based} & \multicolumn{2}{c}{Reconstruction-based} \\ 
        & \textbf{DeSTSeg\cite{zhang2023destseg}}   &\textbf{SimpleNet\cite{liu2023simplenet}} & \textbf{DRÆM\cite{zavrtanik2021draem}} & \textbf{UniAD\cite{you2022unified}}  & \textbf{RD4AD\cite{deng2022rd4ad}} & \textbf{Dinomaly\cite{guo2024dinomaly}}  \\
        \hline      
        
        MTD &95.7	/ \blue{30.2}	/ 67.8	&57.4	/ \pzo6.0	/ 24.9	&79.0	/ 16.2	/ 46.5	&\red{98.8}	/ 27.9	/ \red{74.7}	&96.4	/ 27.6	/ \blue{70.7}	&\blue{98.5}	/ \red{44.6}	/ 68.3\\
        
        STEEL &\blue{65.7}	/ 20.5	/ 24.8		&45.8		/ \pzo3.4		/ 17.2	    	& 55.6		/ 12.8		/ 21.6		& 54.8		/ \blue{24.5}		/ 38.7		&57.1		/ 14.0		/ \red{46.4}		&\red{66.4}	/ \red{31.6} / \blue{43.5}\\
        
        KolektorSDD &77.7		/ \blue{19.6}	/ 49.4	        	&56.7	/ \pzo0.8	/ 23.7	&61.1		/ \pzo2.0		/ 43.6		& \blue{83.3}		/\pzo7.8		/ 47.8		&76.4		/ 12.1		/ \blue{66.8}		&\red{95.3}	/ \red{23.0}  / \red{94.4}\\

        KolektorSDD2 &95.2	/ \red{71.4}	/ 89.9	&54.2	/ \pzo1.0	/ 15.4	&79.5	/ 38.9	/ 58.2	&\blue{96.1}	/ 46.0	 / \blue{94.2}	&\red{96.2}	/ 49.9	/ \red{94.5}	&95.2	/ \blue{70.1}	/ 91.6\\
                
        Casting\_C1 &61.0		/ \pzo0.4		/ 16.7	        	&49.1		/ \pzo0.4		/ 22.6	&\red{79.4}	/ \pzo0.9		/ 41.1		& 51.8		/ \pzo0.4		/ 11.3		&46.1		/ \blue{\pzo2.1}		/ \blue{43.4}		&\blue{75.0}   /  \red{\pzo 3.7} /  \red{43.7}\\

        Casting\_C2 &50.9		/ \pzo1.1		/ \blue{47.6}	        	&49.1		/ \pzo0.3		/ 19.1	&  58.7	/ \pzo1.0		/ 46.9		& 50.6		/ \pzo0.7		/ 34.3		&\red{64.9}		/ \blue{\pzo1.9}	/ \red{62.3}		&\blue{58.8}   /    \red{\pzo 2.6} /    46.9\\

        Casting\_C3 &\blue{72.2}		/ \pzo0.4		/ 31.3	        	&58.5		/ \pzo0.1		/ 12.4	&\red{72.9}	/ \pzo0.5		/ \blue{39.5}		& 50.5		/ \pzo0.3		/ 22.2		&62.6		/    \red{\pzo0.7}		/  \red{50.6}		&67.2   /  \blue{\pzo 0.6} / 
        26.8\\
        
        \hline
        Average &\blue{74.1}	/ \blue{20.5}	/ 46.8    	&53.0	/ \pzo1.7	/ 19.3 	&69.5	/ 10.3	/ 42.5	& 69.4 / 15.4 / 46.2	&71.4	/ 15.5	/ \red{62.1}  &    \red{79.5}	/ \red{25.2}	/ \blue{59.3}	\\        
        \hline
        \end{tabular}
        }
    }
    \vspace{-1.5em}
    \label{tab:iad}
\end{table*}
\subsection{Evaluation Metric}
%EfficientAD中的Latency和Throughput来度量计算成本
The Area Under the Receiver Operating Characteristic curve (AUROC) is the most widely used metric for image-level anomaly detection. However, due to the extreme class imbalance between normal and abnormal pixels in IAD tasks, pixel-level anomaly localization is evaluated using the Area Under the Per-Region Overlap (AUPRO) and Average Precision (AP) metrics~\cite{bergmann2019mvtec}.
%
% \subsection{Standard Training Configuration}
%
%
\subsection{Comparisons with Multi-Class IAD Benchmarks}
%包含定量和定性的分析
%定量实验：多类别设置下 与其他数据集上平均值的不同方法的比较结果----不弄了吧，核心不是为了它；
%         多类别设置下 各个类别上不同对比方法的比较结果；
%定性实验：多类别设置下，不同类别，不同方法的比较结果；失败的案例分析；
%
We comprehensively compare the performance of our HSS-IAD with some popular IAD benchmarks (e.g., MVTec-AD~\cite{bergmann2019mvtec}, VisA~\cite{zou2022visa} Real-IAD~\cite{wang2024real}, BTAD~\cite{mishra2021btad} and MPDD~\cite{jezek2021mpdd}). MVTec-AD is a widely used dataset for the IAD task. The dataset consists of 3,629 normal images for training and a test set of 1,725 images. VisA~\cite{zou2022visa} features 12 different object categories, which contains 8,659 normal images for training and 2,162 images for evaluation. Real-IAD~\cite{wang2024real}, recently released as a large-scale, multi-view anomaly detection dataset, encompasses 30 distinct object categories. BTAD (beanTech Anomaly Detection)~\cite{mishra2021btad} contains 2830 real-world images of 3 industrial products, while MPDD (Metal Parts Defect Detection)~\cite{jezek2021mpdd} comprises over 1346 images across 6 categories with pixel-precise defect annotation masks. Collectively, the BTAD and MPDD datasets serve as robust, real-world benchmarks for IAD.

% Under the multi-class IAD setting, we primarily select methods from three categories for performance comparison: embedding-based IAD approaches, such as DeSTSeg~\cite{zhang2023destseg}
% \footnote{\href{https://github.com/apple/ml-destseg}{https://github.com/apple/ml-destseg}}
% and SimpleNet~\cite{liu2023simplenet}\footnote{\href{https://github.com/DonaldRR/SimpleNet}{https://github.com/DonaldRR/SimpleNet}}; data augmentation–based techniques, including DRÆM~\cite{zavrtanik2021draem}\footnote{\href{https://github.com/VitjanZ/DRAEM}{https://github.com/VitjanZ/DRAEM}} and UniAD~\cite{you2022unified}\footnote{\href{https://github.com/zhiyuanyou/UniAD}{https://github.com/zhiyuanyou/UniAD}}; and reconstruction-based methods, such as RD4AD~\cite{deng2022rd4ad}\footnote{\href{https://github.com/hq-deng/RD4AD}{https://github.com/hq-deng/RD4AD}} and Dinomaly~\cite{guo2024dinomaly}\footnote{\href{https://github.com/guojiajeremy/Dinomaly}{https://github.com/guojiajeremy/Dinomaly}}.
%
Under the multi-class IAD setting, we primarily select methods from three categories for performance comparison: embedding-based IAD approaches, such as DeSTSeg~\cite{zhang2023destseg}
and SimpleNet~\cite{liu2023simplenet}; data augmentation–based techniques, including DRÆM~\cite{zavrtanik2021draem} and UniAD~\cite{you2022unified}; and reconstruction-based methods, such as RD4AD~\cite{deng2022rd4ad} and Dinomaly~\cite{guo2024dinomaly}.
DeSTSeg~\cite{zhang2023destseg} employs a student network trained to denoise feature representations by utilizing both randomly corrupted and normal images. In contrast, SimpleNet~\cite{liu2023simplenet} generates anomalous features by introducing noise to normal features and carefully calibrates the noise scale to maintain a tightly bounded normal feature space. DRÆM~\cite{zavrtanik2021draem} employs a reconstruction-based anomaly embedding model to establish a decision boundary that distinguishes between normal and anomalous instances. UniAD~\cite{you2022unified} prevents the model from falling into an “identical shortcut” by utilizing a neighborhood-masked attention module and a feature jittering strategy.
RD4AD~\cite{deng2022rd4ad} enhances the teacher-student (T-S) model's response to anomalies by leveraging a reverse distillation paradigm.
Dinomaly~\cite{guo2024dinomaly} introduces a minimalist reconstruction-based anomaly detection framework aimed at bridging the performance gap under both multi-class and class-separated settings.
% DiAD~\cite{he2024diad} proposes a diffusion-based framework that reconstructs anomalous regions while preserving the semantic information of the original image.
We reproduce all methods using the official codes.
In our experiments, all hyperparameters, such as preprocessing, batch size, and learning rate, are kept consistent with the official implementation.

The results of all methods on MVTec-AD, VisA, Real-IAD, BTAD, MPDD and HSS-IAD are presented in Table~\ref{tab:compare_other_datasets}. 1) We observe a significant performance drop from MVTec (94.0\% in I-AUROC and 83.2\% in P-AUPRO) to our HSS-IAD (67.0\% in I-AUROC and 43.8\% in P-AUPRO). This suggests that the proposed HSS-IAD dataset is more challenging for anomaly detection and localization than existing datasets. 2) Evaluating different methods on the existing datasets is difficult, as the results are very similar and high-performing. In contrast, on the HSS-IAD dataset, most methods achieve lower I-AUROC and P-AUPRO scores, making it a more effective benchmark for assessing the performance of anomaly detection algorithms.

Table~\ref{tab:miad} presents performance (I-AUROC/P-AP/P-AUPRO) comparison on HSS-IAD under multi-class setting. 1) Although the dataset's subclasses consist of similar metallic-like parts, the performance of image-level anomaly classification varies significantly across categories with differing structures or appearances. Both Casting\_C2 and STEEL achieved maximum I-AUROC scores of only 67.3\% and 63.0\%, respectively, primarily due to the similarity between surface normal features and defect morphologies. For instance, structural features such as threaded holes or through holes are regarded as anomalies because they resemble defect pit shapes. Processing marks that resemble scratches are easily misclassified as defects. Additionally, interference sources like oxide scales, surface slag, and oil stains can lead to model misjudgments, as illustrated in Fig.~\ref{fig_visual}(a).
2) While anomaly detection performance is moderate, anomaly localization performance is generally suboptimal, particularly concerning the P-AP metric. For instance, we observed that the average maximum P-AP and P-AUPRO were only 22.5 and 58.1, respectively. Qualitative analysis reveals that the presence of complex structural features, local interferences, and subtle yet variable surface defect characteristics collectively compound the challenges in surface anomaly detection, as illustrated in Fig.~\ref{fig_visual}(b).
3) Table~\ref{tab:miad} demonstrates that Dinomaly achieves the highest detection and localization performance with values of 77.7/22.5/54.8, followed by DeSTSeg with 73.6/19.8/55.6. In contrast, SimpleNet records the lowest results, 54.3/9.5/20.5. This indicates that adding noise in the latent space is marginally effective in simulating or representing industrial defects. In contrast, leveraging teacher-student model detection mechanisms and the robust representation capabilities of DINOv2 proves highly effective for detecting subtle and confusable surface defects within complex contexts. Furthermore, additional in-depth research efforts are necessary, such as integrating the semantic understanding capabilities of large vision-language models, to further enhance performance.
% Comprehensive experimental results for all types are presented in Table~\ref{tab:hssiad_sp_multi} and \ref{tab:hssiad_px_multi} of the Supplementary Material.
%
%定性和定量结果的分析，并给出挑战难题；
\subsection{Comparisons with Class-Separated IAD}
%包含定量和定性的分析
%定量实验：单类别与其他数据集上平均值的不同方法的比较结果；单类别上各个类别的不同对比方法的比较结果；
%定性实验：单类别设置下，不同类别，不同方法的比较结果；失败的案例分析；
%
%%定性和定量结果的分析，并给出挑战难题，以及可能的方案；
%
To illustrate that the dataset supports multiple detection tasks, we evaluated the performance of SoTA methods under class-separated setting. For comparison with performance under the multi-class case, we selected similar unsupervised anomaly detection methods, including DeSTSeg~\cite{zhang2023destseg}, SimpleNet~\cite{liu2023simplenet}, DRÆM~\cite{zavrtanik2021draem}, UniAD~\cite{you2022unified}, RD4AD~\cite{deng2022rd4ad}, and Dinomaly~\cite{guo2024dinomaly}. The main results are reported in Table~\ref{tab:iad}. 
% More detailed experimental results are presented in Table~\ref{tab:hss_iad_separated} and \ref{tab:hssiad_px_separated} of the Supplementary Material.

Under the Class-Separated UAD setting, most SoTA anomaly detection methods (such as Dinomaly~\cite{guo2024dinomaly}, DeSTSeg~\cite{zhang2023destseg}, and RD4AD~\cite{deng2022rd4ad}) show almost similar performance for image-level anomaly classification. For pixel-level anomaly segmentation, Dinomaly~\cite{guo2024dinomaly}, DeSTSeg~\cite{zhang2023destseg}, and RD4AD~\cite{deng2022rd4ad}, which utilize multi-level features when computing anomaly maps, show a clear advantage, as low-level features retain rich spatial location information. Additionally, castings with distinct structural and visual characteristics show markedly different classification and localization performances, underscoring the rationale and necessity of redefining categories according to their surface properties.
\begin{figure}[htbp]
\centerline{\includegraphics[width=1.0\linewidth]{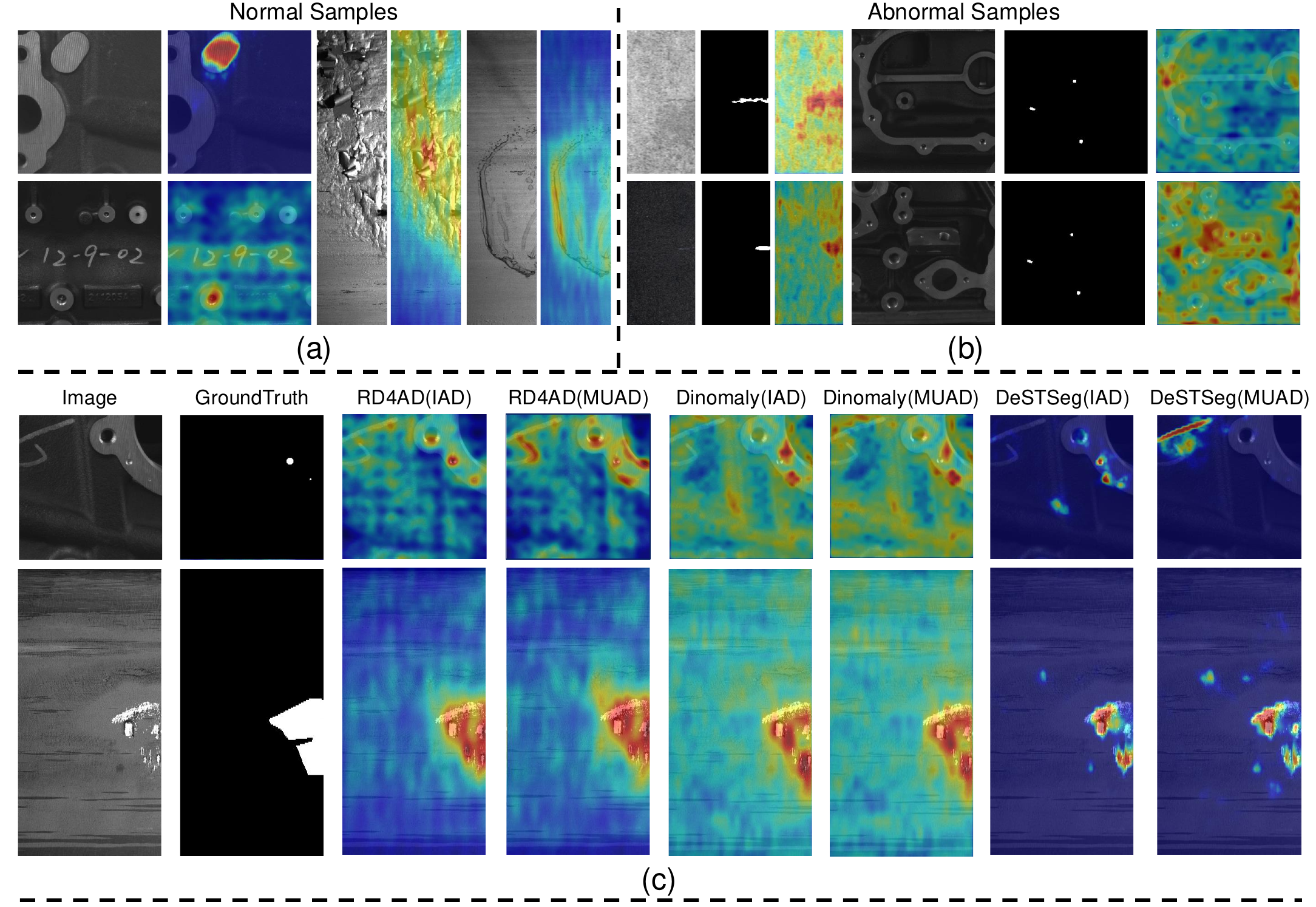}}
\caption{Qualitative results for anomaly localization on HSS-IAD dataset. (a) Misjudged results for normal samples. (b) False positive results for anomaly samples. (e) Visualization results for anomaly localization by different methods.}
\label{fig_visual}
\end{figure}

In the Class-Separated IAD setting, almost all methods demonstrated performance improvements across all metrics, particularly the image (or feature) reconstruction-based approaches RD4AD~\cite{deng2022rd4ad} and Dinomaly~\cite{guo2024dinomaly}, as illustrated in Fig.~\ref{fig_visual}(c). These models typically fall into an identity shortcut, where they return a direct copy of the input regardless of any defects present. As a result, even anomalous samples can be well recovered by the learned model, making them difficult to detect. Moreover, under the unified case where the distribution of normal data is more complex, the identity shortcut problem is exacerbated. Therefore, modeling the distribution of single-class normal data and establishing a compact decision boundary is comparatively straightforward.
Although Dinomaly achieves nearly optimal performance with values of 79.5/25.2/59.3 among all methods, its performance improvements are relatively limited compared to other approaches. This means that IAD algorithms still require further improvements tailored to industrial applications.

\section{CONCLUSION}
% Compared to existing unsupervised anomaly detection datasets, our HSS-IAD distinguishes itself as a heterogeneous same-sort industrial anomaly detection dataset. 
In this paper, we introduce a heterogeneous same-sort industrial anomaly detection (HSS-IAD) dataset. The subclasses in the dataset consist of same sort industrial products, exhibiting variations in structure or appearance. Additionally, the subtle and variable defects of the parts closely resemble the materials themselves. It contains 8,580 metal or magnetic tile surface images, covering 3 objects and 4 texture categories, where all abnormal images include precise anomaly annotations.
To benchmark HSS-IAD, we established two settings, class-separated and multi-class settings, to conduct extensive and comprehensive evaluations using SoTA anomaly detection methods. The experimental findings reveal that HSS-IAD, encompassing intricate contexts and variable structures, holds substantial significance for advancing research in the field of unified anomaly detection for industrial products.

\bibliographystyle{IEEEbib}
\bibliography{HSS-IAD}

\begin{thebibliography}{10}

\bibitem{wang2024casting}
Qishan Wang, Shuyong Gao, Li~Xiong, Aili Liang, Kaidong Jiang, and Wenqiang Zhang,
\newblock ``A casting surface dataset and benchmark for subtle and confusable defect detection in complex contexts,''
\newblock {\em IEEE Sensors Journal}, 2024.

\bibitem{you2022unified}
Zhiyuan You, Lei Cui, Yujun Shen, Kai Yang, Xin Lu, Yu~Zheng, and Xinyi Le,
\newblock ``A unified model for multi-class anomaly detection,''
\newblock {\em Advances in Neural Information Processing Systems}, vol. 35, pp. 4571--4584, 2022.

\bibitem{zhang2023vit}
Jiangning Zhang, Xuhai Chen, Yabiao Wang, Chengjie Wang, Yong Liu, Xiangtai Li, Ming-Hsuan Yang, and Dacheng Tao,
\newblock ``Exploring plain vit reconstruction for multi-class unsupervised anomaly detection,''
\newblock {\em arXiv preprint arXiv:2312.07495}, 2023.

\bibitem{bergmann2019mvtec}
Paul Bergmann, Michael Fauser, David Sattlegger, and Carsten Steger,
\newblock ``Mvtec ad--a comprehensive real-world dataset for unsupervised anomaly detection,''
\newblock in {\em Proceedings of the IEEE/CVF conference on computer vision and pattern recognition}, 2019, pp. 9592--9600.

\bibitem{wang2024real}
Chengjie Wang, Wenbing Zhu, Bin-Bin Gao, Zhenye Gan, Jiangning Zhang, Zhihao Gu, Shuguang Qian, Mingang Chen, and Lizhuang Ma,
\newblock ``Real-iad: A real-world multi-view dataset for benchmarking versatile industrial anomaly detection,''
\newblock in {\em Proceedings of the IEEE/CVF Conference on Computer Vision and Pattern Recognition}, 2024, pp. 22883--22892.

\bibitem{bergmann2022MVTec_LOCO}
Paul Bergmann, Kilian Batzner, Michael Fauser, David Sattlegger, and Carsten Steger,
\newblock ``Beyond dents and scratches: Logical constraints in unsupervised anomaly detection and localization,''
\newblock {\em International Journal of Computer Vision}, vol. 130, no. 4, pp. 947--969, 2022.

\bibitem{survey_ad}
Jiaqi Liu, Guoyang Xie, Jingbao Wang, Shangnian Li, Chengjie Wang, Feng Zheng, and Yaochu Jin,
\newblock ``Deep industrial image anomaly detection: A survey,''
\newblock {\em arXiv preprint arXiv:2301.11514}, 2023.

\bibitem{zhang2023destseg}
Xuan Zhang, Shiyu Li, Xi~Li, Ping Huang, Jiulong Shan, and Ting Chen,
\newblock ``Destseg: Segmentation guided denoising student-teacher for anomaly detection,''
\newblock in {\em Proceedings of the IEEE/CVF Conference on Computer Vision and Pattern Recognition}, 2023, pp. 3914--3923.

\bibitem{zavrtanik2021draem}
Vitjan Zavrtanik, Matej Kristan, and Danijel Sko{\v{c}}aj,
\newblock ``Draem-a discriminatively trained reconstruction embedding for surface anomaly detection,''
\newblock in {\em Proceedings of the IEEE/CVF international conference on computer vision}, 2021, pp. 8330--8339.

\bibitem{qin2022DIS}
Xuebin Qin, Hang Dai, Xiaobin Hu, Deng-Ping Fan, Ling Shao, and Luc Van~Gool,
\newblock ``Highly accurate dichotomous image segmentation,''
\newblock in {\em European Conference on Computer Vision}. Springer, 2022, pp. 38--56.

\bibitem{tabernik2020KolektorSDD}
Domen Tabernik, Samo {\v{S}}ela, Jure Skvar{\v{c}}, and Danijel Sko{\v{c}}aj,
\newblock ``Segmentation-based deep-learning approach for surface-defect detection,''
\newblock {\em Journal of Intelligent Manufacturing}, vol. 31, no. 3, pp. 759--776, 2020.

\bibitem{bovzivc2021ksdd2}
Jakob Bo{\v{z}}i{\v{c}}, Domen Tabernik, and Danijel Sko{\v{c}}aj,
\newblock ``Mixed supervision for surface-defect detection: From weakly to fully supervised learning,''
\newblock {\em Computers in Industry}, vol. 129, pp. 103459, 2021.

\bibitem{huang2020MTD}
Yibin Huang, Congying Qiu, and Kui Yuan,
\newblock ``Surface defect saliency of magnetic tile,''
\newblock {\em The Visual Computer}, vol. 36, no. 1, pp. 85--96, 2020.

\bibitem{severstal-steel-defect-detectionSTEEL}
Alexey Grishin, BorisV, iBardintsev, inversion, and Oleg,
\newblock ``Severstal: Steel defect detection,'' \url{https://kaggle.com/competitions/severstal-steel-defect-detection}, 2019,
\newblock Kaggle.

\bibitem{jezek2021mpdd}
Stepan Jezek, Martin Jonak, Radim Burget, Pavel Dvorak, and Milos Skotak,
\newblock ``Deep learning-based defect detection of metal parts: evaluating current methods in complex conditions,''
\newblock in {\em 2021 13th International congress on ultra modern telecommunications and control systems and workshops (ICUMT)}. IEEE, 2021, pp. 66--71.

\bibitem{zou2022visa}
Yang Zou, Jongheon Jeong, Latha Pemula, Dongqing Zhang, and Onkar Dabeer,
\newblock ``Spot-the-difference self-supervised pre-training for anomaly detection and segmentation,''
\newblock in {\em European Conference on Computer Vision}. Springer, 2022, pp. 392--408.

\bibitem{mishra2021btad}
Pankaj Mishra, Riccardo Verk, Daniele Fornasier, Claudio Piciarelli, and Gian~Luca Foresti,
\newblock ``Vt-adl: A vision transformer network for image anomaly detection and localization,''
\newblock in {\em 2021 IEEE 30th International Symposium on Industrial Electronics (ISIE)}. IEEE, 2021, pp. 01--06.

\bibitem{liu2023simplenet}
Zhikang Liu, Yiming Zhou, Yuansheng Xu, and Zilei Wang,
\newblock ``Simplenet: A simple network for image anomaly detection and localization,''
\newblock in {\em Proceedings of the IEEE/CVF Conference on Computer Vision and Pattern Recognition}, 2023, pp. 20402--20411.

\bibitem{deng2022rd4ad}
Hanqiu Deng and Xingyu Li,
\newblock ``Anomaly detection via reverse distillation from one-class embedding,''
\newblock in {\em Proceedings of the IEEE/CVF conference on computer vision and pattern recognition}, 2022, pp. 9737--9746.

\bibitem{guo2024dinomaly}
Jia Guo, Shuai Lu, Weihang Zhang, Fang Chen, Hongen Liao, and Huiqi Li,
\newblock ``Dinomaly: The less is more philosophy in multi-class unsupervised anomaly detection,''
\newblock {\em arXiv preprint arXiv:2405.14325}, 2024.

\end{thebibliography}

% \clearpage
% \setcounter{page}{1}
% \input{./sec/6_suppl}

\vspace{12pt}
\color{red}
\end{document}